\pdfoutput=1
\documentclass[11pt]{article}
\usepackage[utf8]{inputenc}
\usepackage{eacl2014}
\makeatletter
\newcommand{\@BIBLABEL}{\@emptybiblabel}
\newcommand{\@emptybiblabel}[1]{}
\makeatother
\usepackage{times}
\usepackage{url}
\usepackage{latexsym}
\usepackage{mathtools}
\usepackage{graphicx}
\usepackage{subfig}
\usepackage{multirow}
\usepackage{fancyhdr} 
\usepackage{hyperref}

\title{Translation Memory Retrieval Methods}

\author{Michael Bloodgood \\
  Center for Advanced Study of Language \\
  University of Maryland \\
  College Park, MD 20742 USA \\
  {\tt meb@umd.edu} \\\And
  Benjamin Strauss \\
  Center for Advanced Study of Language \\
  University of Maryland \\
  College Park, MD 20742 USA \\
  {\tt bstrauss@umd.edu} \\}

\date{}

\begin{document}

\thispagestyle{fancy}

\maketitle
\begin{abstract}
Translation Memory (TM) systems are one of the most widely used translation technologies. 
An important part of TM systems is the matching algorithm that determines what translations get retrieved from the bank of available 
translations to assist the human translator. 
Although detailed accounts of the matching algorithms used in commercial systems can't be found in the literature, 
it is widely believed that edit distance algorithms are used. 
This paper investigates and evaluates the use of several matching algorithms, 
including the edit distance algorithm that is believed to be at the heart of most modern commercial TM systems. 
This paper presents results showing how well various matching algorithms correlate with human judgments 
of helpfulness (collected via crowdsourcing with Amazon's Mechanical Turk). 
A new algorithm based on 
weighted n-gram precision that can be adjusted for translator length preferences 
consistently returns translations judged to be most helpful by translators for multiple domains and 
language pairs. 

\end{abstract}

\section{Introduction} \label{intro}

The most widely used computer-assisted translation (CAT) tool for professional translation of specialized text is translation memory (TM) 
technology \cite{christensen2010}.  
TM consists of a database of previously translated material, referred to as the TM vault or the TM bank (TMB in the rest of this paper). 
When a translator is translating a new sentence, the TMB is consulted to see if a similar sentence has already been translated and if so, the 
most similar previous translation is retrieved from the bank to help the translator. 
The main conceptions of TM technology occurred in the late 1970s and early 1980s \cite{arthern1978,kay1980,melby1981}. 
TM has been widely used since the late 1990s and continues to be widely used today \cite{bowker2008,christensen2010,garcia2007,somers2003}. 

There are a lot of factors that determine how helpful TM technology will be in practice. Some of these include: quality of the interface,
speed of the back-end database lookups, speed of network connectivity for distributed setups, and the comfort of the translator with 
using the technology.  
A fundamentally important factor that determines how helpful TM technology will be in practice is how well the TM bank of previously
translated materials matches up with the workload materials to be translated. It is necessary that there be a high level of match for the TM
technology to be most helpful. However, having a high level of match is not sufficient. One also needs a successful method for retrieving
the useful translations from the (potentially large) TM bank. 

TM similarity metrics are used for both evaluating the expected helpfulness of previous translations for new workload translations and the
metrics also directly determine what translations get provided to the translator during translation of new materials. 
Thus, the algorithms that compute the TM similarity metrics are not only important, but they are doubly important. 

The retrieval algorithm used by commercial TM systems is typically not disclosed \cite{koehn2010,simard2012,whyman1999}. 
However, the best-performing method used in current systems is widely believed to be based on edit distance 
\cite{baldwin2000,simard2012,whyman1999,koehn2010,christensen2010,mandreoli2006,he2010}.
Recently \newcite{simard2012} have experimented with using MT (machine translation) evaluation metrics as TM fuzzy match, or similarity, 
algorithms. 
A limitation of the work of \cite{simard2012} was that the evaluation of the performance of the TM similarity algorithms 
was also conducted using the same MT evaluation metrics.
\newcite{simard2012} concluded that their evaluation of TM similarity functions was biased since whichever MT evaluation metric was 
used as the TM similarity function was also likely to obtain the best score under that evaluation metric.

The current paper explores various TM fuzzy match algorithms ranging from simple baselines to the widely used edit distance to 
new methods. The evaluations of the TM fuzzy match algorithms use 
human judgments of helpfulness. An algorithm based on weighted n-gram 
precision consistently returns translations judged to be most helpful by translators for multiple domains and language pairs. 
In addition to being able to retrieve useful translations from the TM bank, the fuzzy match scores ought to be indicative of how helpful a
translation can be expected to be. Many translators find it counter-productive to use TM when the best-matching translation from the TM is
not similar to the workload material to be translated. Thus, many commercial TM products offer translators the opportunity to set a fuzzy
match score threshold so that only translations with scores above the threshold will ever be returned. It seems to be a widely used practice
to set the threshold at 70\% but again it remains something of a black-box as to why 70\% ought to be the setting.
The current paper uncovers what expectations of helpfulness can be given for different threshold settings for various fuzzy
match algorithms. 

The rest of this paper is organized as follows. Section~\ref{metrics} presents the TM similarity metrics that will be explored; 
section~\ref{setup} presents our experimental setup; section~\ref{results} presents and analyzes results; 
and section~\ref{conclusions} concludes.

\section{Translation Memory Similarity Metrics} \label{metrics}

In this section we define the methods for measuring TM similarity for which experimental results are reported in section~\ref{results}.
All of the metrics compute scores between 0 and 1, with higher scores indicating better matches. 
All of the metrics take two inputs: $M$ and $C$, where $M$ is a workload sentence from the MTBT (Material To Be Translated) and $C$ is the source language side of 
a candidate pre-existing translation from the TM bank. 
The metrics range from simple baselines to the surmised current industrial standard to new methods.

\subsection{Percent Match}

Perhaps the simplest metric one could conceive of being useful for TM similarity matching is percent match (PM), 
the percent of tokens in the MTBT segment found in the source language side of the candidate translation pair from the TM bank.

Formally,
\begin{equation} \label{PM}
PM(M,C) = \frac{|M_{unigrams} \bigcap C_{unigrams}|}{|M_{unigrams}|},
\end{equation}
where $M$ is the sentence from the MTBT that is to be translated, 
$C$ is the source language side of the candidate translation from the TM bank,
$M_{unigrams}$ is the set of unigrams in $M$,
and $C_{unigrams}$ is the set of unigrams in $C$. 

\subsection{Weighted Percent Match}

A drawback of PM is that it weights the matching of each unigram in an MTBT segment equally, however, it is not the case that the value of assistance to
the translator is equal for each unigram of the MTBT segment. The parts that are most valuable to the translator are the parts that he/she does not
already know how to translate. Weighted percent match (WPM) uses inverse document frequency (IDF) as a proxy for trying to weight words based on how much
value their translations are expected to provide to translators. The use of IDF-based weighting is motivated by the assumption that 
common words that permeate throughout the language will be easy for translators to translate but words that occur in relatively rare situations will be 
harder to translate and thus more valuable to match in the TM bank. For our implementation of WPM, each source language sentence in the parallel corpus 
we are experimenting with is treated as a ``document" when computing IDF.

Formally,
\begin{align} \label{WPM}
WP&M(M,C) = \nonumber
\\ &\frac{\sum\limits_{u \in \{M_{unigrams} \bigcap C_{unigrams}\}}^{} idf(u,D)}{\sum\limits_{u \in M_{unigrams}}^{} idf(u,D)},
\end{align}
where $M$, $C$, $M_{unigrams}$, and $C_{unigrams}$ are as defined in Eq.~\ref{PM},
$D$ is the set of all source language sentences in the parallel corpus,
and $idf(x,D) = log(\frac{|D|}{|\{d \in D: x \in d\}|})$.

\subsection{Edit Distance}

A drawback of both the PM and WPM metrics are that they are only considering coverage of the words from the 
workload sentence in the candidate sentence from the TM bank and not taking into account the context of the words. However, words can be translated very differently 
depending on their context. Thus, a TM metric that matches sentences on more than just (weighted) percentage coverage of lexical items can be expected to 
perform better for TM bank evaluation and retrieval. 
Indeed, as was discussed in section~\ref{intro}, it is widely believed that most TM similarity metrics used in existing systems are based on string edit distance.

Our implementation of edit distance \cite{levenshtein1966}, computed on a word level, is similar to the version 
defined in \cite{koehn2010}. 

Formally, our TM metric based on Edit Distance (ED) is defined as
\begin{equation} \label{ED}
ED=max\left(
1-\frac{edit \mbox{-} dist(M,C)}{|M_{unigrams}|},0\right),
\end{equation}
where $M$, $C$, and $M_{unigrams}$ are as defined in Eq.~\ref{PM}, and $edit \mbox{-} dist(M,C)$ is the number of word deletions, 
insertions, and substitutions required to transform $M$ into $C$.

\subsection{N-Gram Precision}

Although ED takes context into account, it does not emphasize local context in matching certain high-value words and phrases as much as metrics that capture n-gram precision between the MTBT workload sentence and candidate source-side sentences from the TMB. 
We note that n-gram precision forms a fundamental subcomputation in the computation of the corpus-level MT evaluation metric 
BLEU score \cite{papineni2002}.   
However, although TM fuzzy matching metrics are related to automated MT evaluation metrics, there are some important differences. 
Perhaps the most important is that TM fuzzy matching has to be able to operate at a sentence-to-sentence level whereas automated MT evaluation metrics
such as BLEU score are intended to operate over a whole corpus. 
Accordingly, we make modifications to how we use n-gram precision for the purpose of TM matching than how we use it when we compute BLEU scores. 
The rest of this subsection and the next two subsections describe the innovations we make in adapting the notion of n-gram precision to the TM matching
task.

Our first metric along these lines, N-Gram Precision (NGP), is defined formally as follows:

\begin{equation} \label{NGP}
NGP = \sum_{n=1}^{N} \frac{1}{N}p_{n},
\end{equation}
where the value of $N$ sets the upper bound on the length of n-grams considered\footnote{We used $N=4$ in our experiments.}, and

\begin{align}
&p_n = \nonumber \\
&\frac{|M_{n \mbox{-} grams} \cap C_{n \mbox{-} grams}|}{Z*|M_{n \mbox{-} grams}|+(1-Z)*|C_{n \mbox{-} grams}|}, \label{p_n}
\end{align}
where $M$ and $C$ are as defined in Eq.~\ref{PM},
$M_{n \mbox{-} grams}$ is the set of n-grams in $M$,
$C_{n \mbox{-} grams}$ is the set of n-grams in $C$,
and $Z$ is a user-set parameter that controls how the metric is normalized.\footnote{Note that the n in n-grams is intended to 
be substituted with the corresponding integer. 
Accordingly, for $p_1$, $n=1$ and therefore $M_{n \mbox{-} grams} = M_{1 \mbox{-} grams}$ is the set of unigrams in $M$ and 
$C_{n \mbox{-} grams} = C_{1 \mbox{-} grams}$ is the set of unigrams in $C$;
for $p_2$, $n=2$ and therefore $M_{n \mbox{-} grams} = M_{2 \mbox{-} grams}$ is the set of bigrams in $M$ and 
$C_{n \mbox{-} grams} = C_{2 \mbox{-} grams}$ is the set of bigrams in $C$; and so on.}

As seen by equation~\ref{NGP}, we use an arithmetic mean of precisions instead of the geometric mean that BLEU score uses. 
An arithmetic mean is better than a geometric mean for use in translation memory metrics since translation memory metrics are operating at a segment level 
and not at the aggregate level of an entire test set. 
At the extreme, the geometric mean will be zero if any of the n-gram precisions $p_n$ are zero.
Since large n-gram matches are unlikely on a segment level, using a geometric mean can be a poor method to use for matching on a segment level, as has been described for the
related task of MT evaluation \cite{doddington2002,lavie2004}.  
Additionally, for the related task of MT evaluation at a segment level, \newcite{lavie2004} have found that using an arithmetic mean correlates better with human judgments
than using a geometric mean.

Now we turn to discussing the parameter Z for controlling how the metric is normalized. 
At one extreme, setting Z=1 will correspond to having no penalty on the length of the candidate retrieved from the TMB and leads to getting 
longer translation matches retrieved. 
At the other extreme, setting Z=0 will correspond to a
normalization that penalizes relatively more for length of the retrieved candidate and leads to shorter translation matches being retrieved. 
There is a precision/recall tradeoff in that one wants to retrieve candidates from the TMB that have high recall in the sense of matching what is in the MTBT
sentence yet one also wants the retrieved candidates from the TMB to have high precision in the sense of not having extraneous material not relevant to helping
with the translation of the MTBT sentence. The optimal setting of Z may differ for different scenarios based on factors like the languages, the corpora, and 
translator preference. We believe that for most TM applications there will usually be an asymmetric valuation of precision/recall in that 
recall will be more important since the value of getting a match will be more than the cost of extra material up to a point. Therefore, we believe a Z setting
in between 0.5 and 1.0 will be an optimal default. 
We use Z=0.75 in all of our experiments described in section~\ref{setup} and reported on in section~\ref{results} except for the experiments explicitly showing the impact of changing the Z parameter. 

\subsection{Weighted N-Gram Precision}

Analogous to how we improved PM with WPM, we seek to improve NGP in a similar fashion. 
As can be seen from the numerator of Equation~\ref{p_n}, NGP is weighting the match of all n-grams as uniformly important. 
However, it is not the case that each n-gram is of equal value to the translator.
Similar to WPM, we use IDF as the basis of our proxy for weighting n-grams according to the value their translations are expected to provide to translators. 
Specifically, we define the weight of an n-gram to be the sum of the IDF values for each constituent unigram that comprises the n-gram. 

Accordingly, we formally define method Weighted N-Gram Precision (WNGP) as follows:

\begin{equation} \label{WNGP}
WNGP=\sum_{n=1}^{N} \frac{1}{N}wp_n,
\end{equation}
where $N$ is as defined in Equation~\ref{NGP}, and 

\begin{align}
&wp_n = \nonumber \\
&\frac{\sum\limits_{i \in \{M_{n \mbox{-} grams} \mbox{ } \cap \mbox{ } C_{n \mbox{-} grams}\}}^{ } \!\!\!\!\!\!\!\!\!\!\!\!\!\!\!\!\!\!\!\!\!\!\!w(i)}  
	{Z \Bigg[\sum\limits_{i \in M_{n \mbox{-} grams}}^{ } \!\!\!\!\!\!\!\!\!w(i) \Bigg] +
	(1-Z) \Bigg[\sum\limits_{i \in C_{n \mbox{-} grams}}^{ } \!\!\!\!\!\!\!\!\!w(i) \Bigg]}, \label{wp_n}
\end{align}
where $Z$, $M_{n \mbox{-} grams}$, and $C_{n \mbox{-} grams}$ are as defined in Equation~\ref{p_n}, and 

\begin{equation} \label{w(i)}
w(i) = \sum\limits_{1 \mbox{-} gram \in i}^{} idf(1 \mbox{-} gram,D),
\end{equation}
where $i$ is an n-gram and $idf(x,D)$ is as defined above for Equation~\ref{WPM}. 

\subsection{Modified Weighted N-gram Precision}

Note that in Equation~\ref{WNGP} each $wp_n$ contributes equally to the average. Modified Weighted N-Gram Precision (MWNGP) improves on 
WNGP by weighting the contribution of each $wp_n$ so that shorter n-grams contribute more than longer n-grams. The intuition is that for
TM settings, getting more high-value shorter n-gram matches at the expense of fewer longer n-gram matches will be more helpful since
translators will get relatively more assistance from seeing new high-value vocabulary. Since the translators already presumably know the rules 
of the language in terms of how to order words correctly, the loss of the longer n-gram matches will be mitigated.  

Formally we define MWNGP as follows:

\begin{equation} \label{MWNGP}
MWNGP = \frac{2^N}{2^N-1} \sum_{n=1}^{N} \frac{1}{2^n}wp_{n},
\end{equation}
where $N$ and $wp_n$ are as they were defined for Equation~\ref{WNGP}.  

\section{Experimental Setup} \label{setup}

We performed experiments on two corpora from two different technical domains with two language pairs, 
French-English and Chinese-English. 
Subsection~\ref{corpora} discusses the specifics of the corpora and the processing we performed. 
Subsection~\ref{human} discusses the specifics of our human evaluations of how helpful retrieved segments are for translation.

\subsection{Corpora} \label{corpora}

For Chinese-English experiments, we used the OpenOffice3 (OO3) parallel corpus \cite{tiedemann2009}, which is OO3 computer office productivity software documentation. 
For French-English experiments, we used the EMEA parallel corpus \cite{tiedemann2009}, which are medical documents from the European Medecines Agency.
The corpora were produced by a suite of automated tools as described in \cite{tiedemann2009} and come sentence-aligned. 

The first step in our experiments was to preprocess the corpora. For Chinese corpora we tokenize each sentence using the Stanford Chinese 
Word Segmenter \cite{tseng2005} with the Chinese Penn Treebank standard \cite{xia2000}. 
For all corpora we remove all segments that have fewer than 5 tokens or more than 100 tokens. We call the resulting set the valid segments.
For the purpose of computing match statistics, for French corpora we remove all punctuation, numbers, and scientific symbols; we case-normalize the text and stem the
corpus using the NLTK French snowball stemmer.
For the purpose of computing match statistics, for Chinese corpora we remove all but valid tokens. Valid tokens must include at least one Chinese character. 
A Chinese character is defined as a character in the Unicode range 0x4E00-0x9FFF or 0x4000-0x4DFF or 0xF900-0xFAFF. 
The rationale for removing these various tokens from consideration for the purpose of computing match statistics is that translation of numbers (when they're written as
Arabic numerals), punctuation, etc. is the
same across these languages and therefore we don't want them influencing the match computations. But once a translation is selected as being most helpful for
translation, the original version (that still contains all the numbers, punctuation, case markings, etc.) is the version that is brought back 
and displayed to the translator. 

For the TM simulation experiments, we randomly sampled 400 translations from the OO3 corpus and pretended that the Chinese sides of those 400 translations constitute the
workload Chinese MTBT. 
From the rest of the corpus we randomly sampled 10,000 translations and pretended that that set of 10,000 translations constitutes the Chinese-English TMB. 
We also did similar sampling from the EMEA corpus of a workload French MTBT of size 300 and a French-English TMB of size 10,000.

After the preprocessing and selection of the TMB and MTBT, we found the best-matching segment from the TMB for each MTBT segment according to each TM retrieval metric
defined in section~\ref{metrics}.\footnote{If more than one segment from the TMB was tied for being the highest-scoring segment, the segment located first in the TMB was
considered to be the best-matching segment.} The resulting sets of (MTBT segment,best-matching TMB segment) pairs formed the inputs on which we conducted our evaluations of
the performance of the various TM retrieval metrics. 

\subsection{Human Evaluations} \label{human}

To conduct evaluations of how helpful the translations retrieved by the various TM retrieval metrics would be for translating the MTBT 
segments, we used Amazon Mechanical Turk, which has been used productively in the past for related work in the context of 
machine translation \cite{bloodgood2010a,bloodgood2010b,callison-burch2009}. 

For each (MTBT segment,best-matching TMB segment) pair generated as discussed in subsection~\ref{corpora}, we collected judgments 
from Turkers (i.e., the workers on MTurk) on how helpful the TMB translation would be for translating the MTBT segment on a 5-point scale. 
The 5-point scale was as follows:

\begin{itemize}
  \item 5 = Extremely helpful. The sample is so similar that with trivial modifications I can do the translation.
  \item 4 = Very helpful. The sample included a large amount of useful words or phrases and/or some extremely useful words or phrases that overlapped with the MTBT.
  \item 3 = Helpful. The sample included some useful words or phrases that made translating the MTBT easier.
  \item 2 = Slightly helpful. The sample contained only a small number of useful words or phrases to help with translating the MTBT.
  \item 1 = Not helpful or detrimental. The sample would not be helpful at all or it might even be harmful for translating the MTBT.
\end{itemize}

After a worker rated a (MTBT segment,TMB segment) pair the worker was then required to give an explanation for their rating. 
These explanations proved quite helpful as discussed in section~\ref{results}. 
For each (MTBT segment,TMB segment) pair, we collected judgments from five different Turkers.
For each (MTBT segment,TMB segment) pair these five judgments were then averaged to form a mean opinion score (MOS) on the helpfulness of the 
retrieved TMB translation for translating the MTBT segment. 
These MOS scores form the basis of our evaluation of the performance of the different TM retrieval metrics. 

\section{Results and Analysis} \label{results}

\subsection{Main Results} \label{mainResults}

Tables \ref{t:OO3Agreement} and \ref{t:EMEAAgreement} show the percent of the time that each pair of metrics agree on the choice of the most helpful TM segment for the Chinese-English OO3 data and the French-English EMEA data, respectively. A main observation to be made is that the choice of metric makes a big difference in the choice of the most helpful TM segment. For example, we can see that the surmised industrial standard ED metric agrees with the new MWNGP metric less than 40\% of the time on both sets of data (35.0\% on Chinese-English OO3 and 39.3\% on French-English EMEA data). 

\begin{table}
\resizebox{\columnwidth}{!}{
\begin{tabular}{r r r r r r r r}
metric & PM & WPM & ED & NGP & WNGP & MWNGP \\
PM & 100.0 & 69.5 & 23.0 & 32.0 & 31.5 & 35.5 \\
WPM & 69.5 & 100.0 & 25.8 & 37.0 & 39.0 & 44.2 \\
ED & 23.0 & 25.8 & 100.0 & 41.5 & 35.8 & 35.0 \\
NGP & 32.0 & 37.0 & 41.5 & 100.0 & 77.8 & 67.0 \\
WNGP & 31.5 & 39.0 & 35.8 & 77.8 & 100.0 & 81.2 \\
MWNGP & 35.5 & 44.2 & 35.0 & 67.0 & 81.2 & 100.0 \\  
\end{tabular}
}
\caption{OO3 Chinese-English: The percent of the time that each pair of metrics agree on the most helpful TM segment}
\label{t:OO3Agreement}
\end{table}

\begin{table}
\resizebox{\columnwidth}{!}{
\begin{tabular}{r r r r r r r r}
metric & PM & WPM & ED & NGP & WNGP & MWNGP \\
PM & 100.0 & 64.7 & 30.3 & 40.3 & 38.3 & 41.3 \\
WPM & 64.7 & 100.0 & 32.0 & 46.3 & 47.0 & 54.3 \\
ED & 30.3 & 32.0 & 100.0 & 42.3 & 40.3 & 39.3 \\
NGP & 40.3 & 46.3 & 42.3 & 100.0 & 76.3 & 67.7 \\
WNGP & 38.3 & 47.0 & 40.3 & 76.3 & 100.0 & 81.3 \\
MWNGP & 41.3 & 54.3 & 39.3 & 67.7 & 81.3 & 100.0 \\
\end{tabular}
}
\caption{EMEA French-English: The percent of the time that each pair of metrics agree on the most helpful TM segment}
\label{t:EMEAAgreement}
\vspace{-.2cm}
\end{table}

\begin{figure}
\includegraphics[keepaspectratio=true, width=\columnwidth]{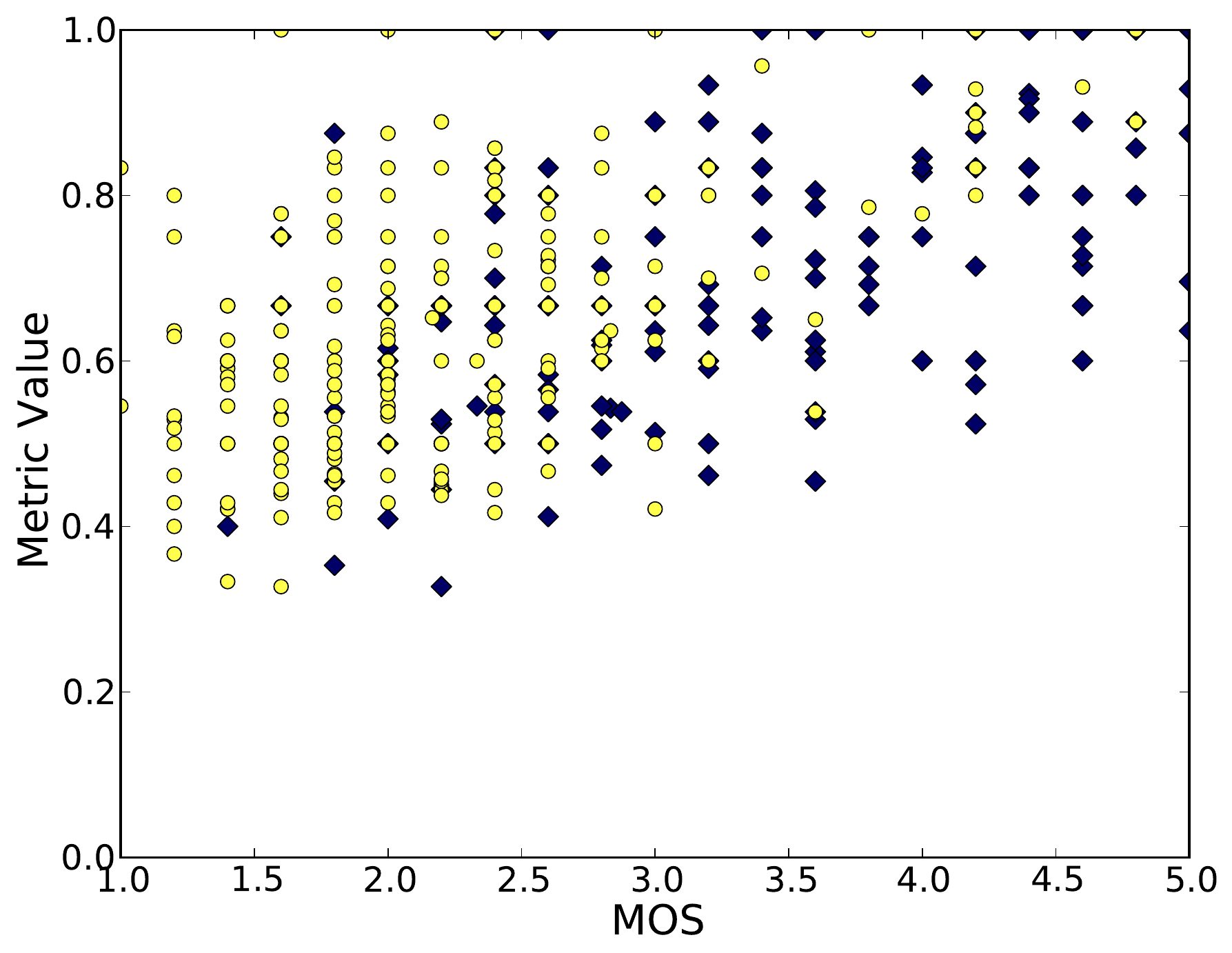}
\caption{OO3 PM scatterplot}
\label{f:PMScatterplotOO3}
\vspace{-.5cm}
\end{figure}

\begin{figure}
\includegraphics[keepaspectratio=true, width=\columnwidth]{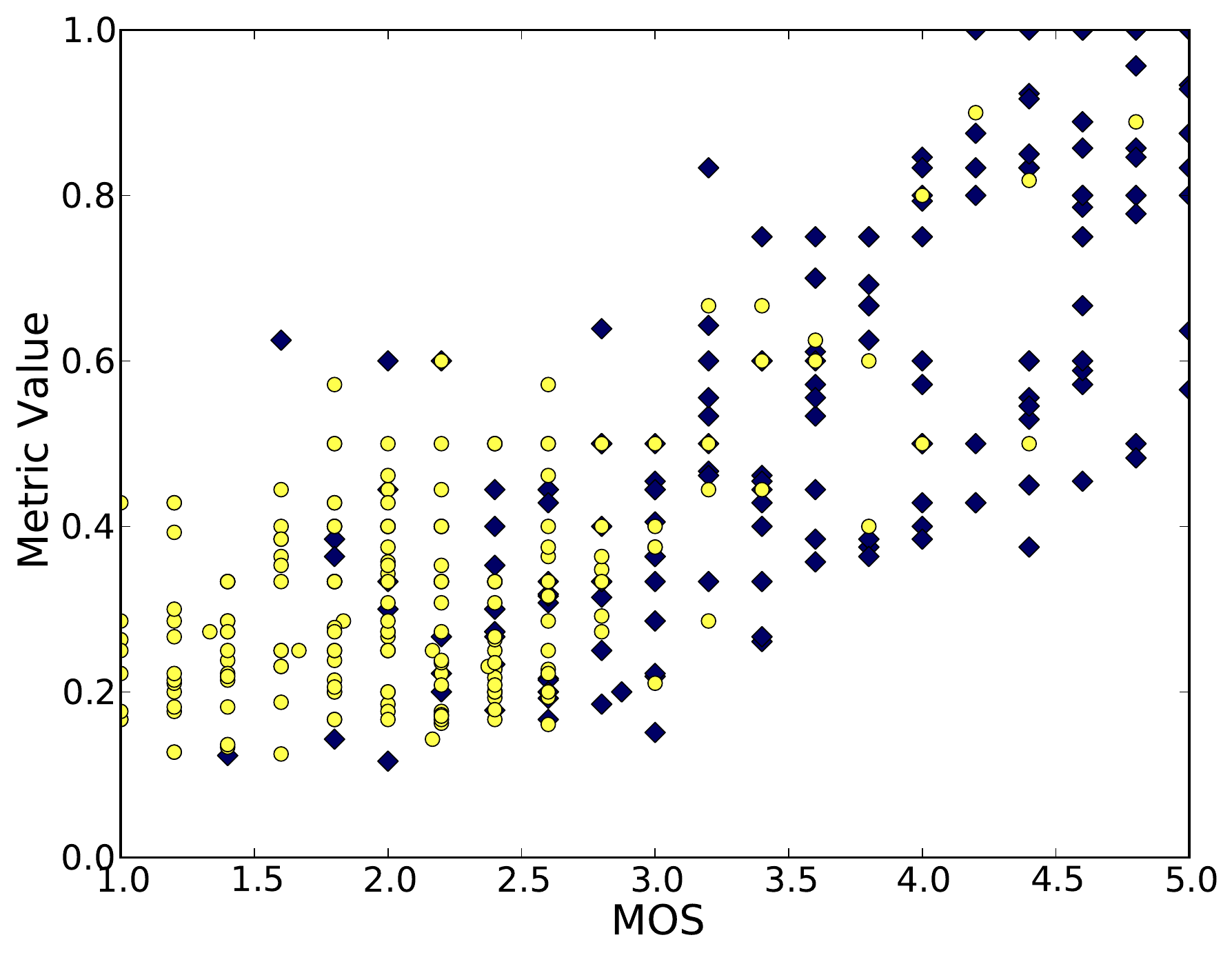}
\caption{OO3 ED scatterplot}
\label{f:EDScatterplotOO3}
\vspace{-.5cm}
\end{figure}

\begin{figure}
\includegraphics[keepaspectratio=true, width=\columnwidth]{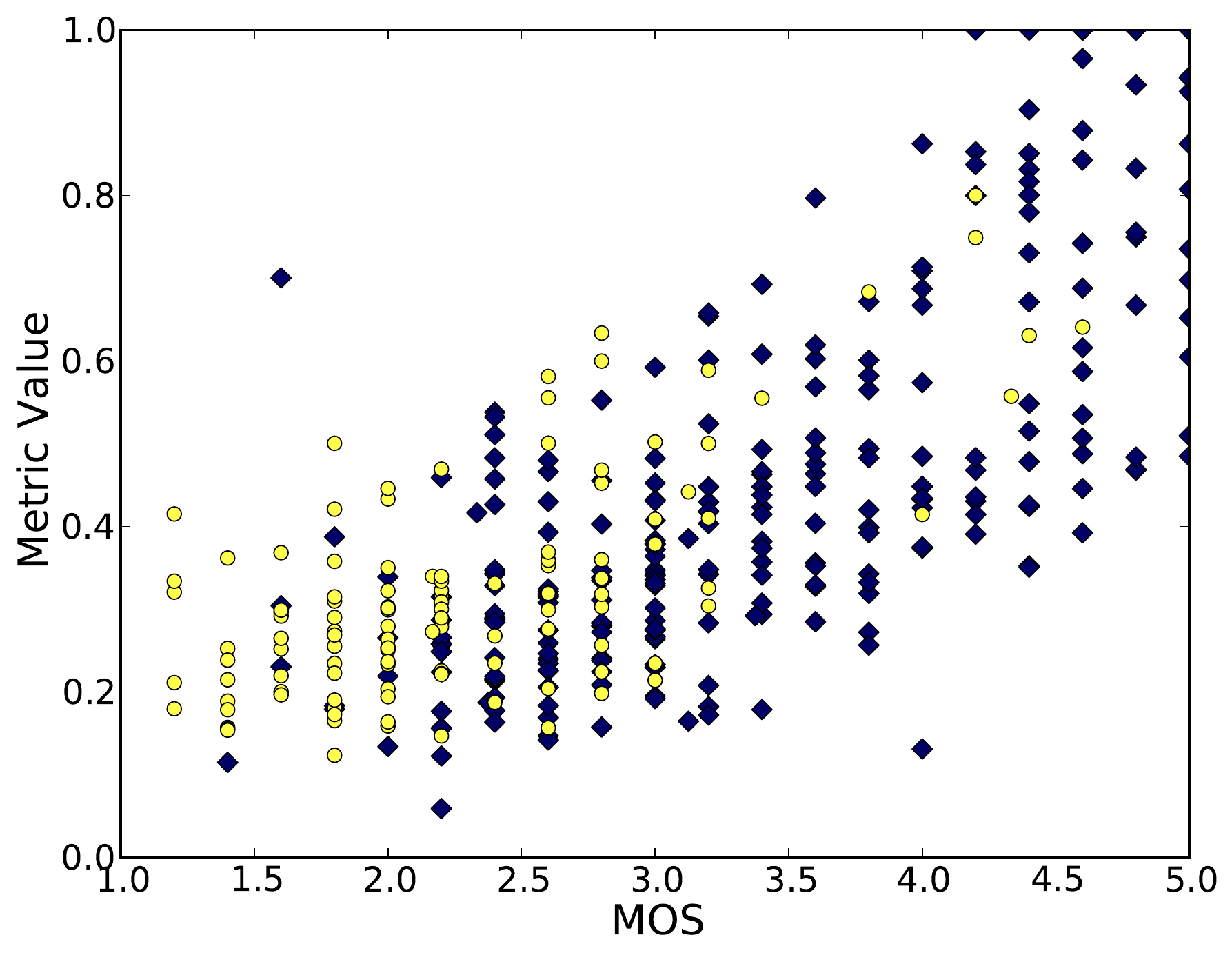}
\caption{OO3 MWNGP scatterplot}
\label{f:MWNGPScatterplotOO3}
\vspace{-.5cm}
\end{figure}

Tables \ref{t:OO3FoundBest} and \ref{t:EMEAFoundBest} show the number of times each metric found the TM segment that
the Turkers judged to be the most helpful out of all the TM segments retrieved by all of the different metrics. From these tables one can see that the MWNGP method consistently retrieves the best TM segment more often than each of the other metrics. Scatterplots showing the exact performance on every MTBT segment of the OO3 dataset for various metrics are shown in Figures \ref{f:PMScatterplotOO3}, \ref{f:EDScatterplotOO3}, and \ref{f:MWNGPScatterplotOO3}.  To conserve space, scatterplots are only shown for metrics PM (baseline metric), ED (strong surmised industrial standard metric), and MWNGP (new highest-performing metric). For each MTBT segment, there is a point in the scatterplot. The y-coordinate is the value assigned by the TM metric to the segment retrieved from the TM bank and the x-coordinate is the MOS of the five Turkers on how helpful the retrieved TM segment would be for translating the MTBT segment. A point is depicted as a dark blue diamond if none of the other metrics retrieved a segment with higher MOS judgment for that MTBT segment. A point is depicted as a yellow circle if another metric retrieved a different segment from the TM bank for that MTBT segment that had a higher MOS. 

\begin{table}
\begin{tabular}{r r r}
Metric  &  Found Best  &  Total MTBT Segments \\
PM &  178  &  400 \\
WPM &  200  &  400 \\
ED &  193  &  400 \\
NGP &  251  &  400 \\
WNGP &  271  &  400 \\
MWNGP &  282  &  400 \\
\end{tabular}
\caption{OO3 Chinese-English: The number of times that each metric found the most helpful TM segment (possibly tied).}
\label{t:OO3FoundBest}
\end{table}

\begin{table}
\begin{tabular}{r r r}
Metric  &  Found Best  &  Total MTBT Segments \\
PM &  166  &  300 \\
WPM &  184  &  300 \\
ED &  148  &  300 \\
NGP &  188  &  300 \\
WNGP &  198  &  300 \\
MWNGP &  201  &  300 \\
\end{tabular}
\caption{EMEA French-English: The number of times that each metric found the most helpful TM segment (possibly tied).}
\label{t:EMEAFoundBest}
\vspace{-.2cm}
\end{table}

A main observation from Figure~\ref{f:PMScatterplotOO3} is that PM is failing as 
evidenced by the large number of points in the upper left quadrant. For those points, the metric value is high, indicating that the retrieved 
segment ought to be helpful. However, the MOS is low, indicating that the humans are judging it to not be helpful. 
Figure~\ref{f:EDScatterplotOO3} shows that the ED metric does not suffer from this problem. However, Figure~\ref{f:EDScatterplotOO3} shows 
that ED has another problem, which is a lot of yellow circles in the lower left quadrant. 
Points in the lower left quadrant are not necessarily indicative of a poorly performing metric, depending on the degree of match of the TMB with the MTBT workload. 
If there is nothing available in the TMB that would help with the MTBT, it is appropriate for the metric to assign a low value and the humans to correspondingly 
agree that the retrieved sentence is not helpful. 
However, the fact that so many of ED's points are yellow circles indicates that there were better segments available in the TMB that ED was not able to 
retrieve yet another metric was able to retrieve them. 
Observing the scatterplots for ED and those for MWNGP one can see that both methods have the vast majority of points concentrated 
in the lower left and upper right quadrants, solving the upper left quadrant problem of PM. 
However, MWNGP has a relatively more densely populated upper right quadrant populated with dark blue diamonds than ED does whereas ED has a more 
densely populated lower left quadrant with yellow circles than MWNGP does. These results and trends are consistent across the EMEA French-English dataset so those scatterplots are omitted to conserve space. 

Examining outliers where MWNGP assigns a high metric value yet the Turkers indicated that the translation has low helpfulness such as the point in 
Figure~\ref{f:MWNGPScatterplotOO3} at (1.6,0.70) is informative. Looking only at the source side, it looks like the translation retrieved from the TMB ought 
to be very helpful. The Turkers put in their explanation of their scores that the reason they gave low helpfulness is because the English translation was incorrect.
This highlights that a limitation of MWNGP, and all other TM metrics we're aware of, is that they only consider the source side. 

\subsection{Adjusting for length preferences} \label{varyingParameters}

As discussed in section~\ref{metrics}, the Z parameter can be used to control for length preferences.
Table~\ref{t:Z_Param} shows how the average length, measured by number of tokens of the source side of the translation pairs returned by MWNGP, changes as the Z parameter is changed. 

\begin{table}
\resizebox{\columnwidth}{!}{
\centering
\subfloat[EMEA French-English\label{t:EMEA_Z_Param}]{
\centering
\begin{tabular}{rr}
\bf{Z Value} & \bf{Avg Length}\\
0.00 & 9.9298\\
0.25 & 13.204\\
0.50 & 16.0134\\
0.75 & 19.6355\\
1.00& 27.8829\\
\end{tabular}
}
\subfloat[OO3 Chinese-English\label{t:OO3_Z_Param}]{
\centering
\begin{tabular}{rr}
\bf{Z Value} & \bf{Avg Length}\\
0.00  &  7.2475   \\ 
0.25   &  9.5600   \\ 
0.50   &  11.1250  \\ 
0.75   &  14.1825  \\ 
1.00   &  25.0875  \\ 
\end{tabular}
}
}
\caption{Average TM segment length, measured by number of tokens of the source side of the translation pairs returned by MWNGP, for varying values of the Z parameter}\label{t:Z_Param}
\vspace{-.6cm}
\end{table}

\begin{table*}
\begin{center}
\begin{tabular}{|l|l|} \hline
MTBT    & {\bf French:} Ne pas utiliser durant la gestation et la lactation, car l’ innocuité du \\
& médicament vétérinaire n’ a pas été établie pendant la gestation ou \\
& la lactation. \\ 
        & {\bf English:} Do not use during pregnancy and lactation because the safety of the \\
& veterinary medicinal product has not been established during \\
& pregnancy and lactation.\\ \hline
MWNGP & {\bf French:} Peut être utilisé pendant la gestation et la lactation.\\ 
(Z=0.00) & {\bf English:} Can be used during pregnancy and lactation.\\ \hline
MWNGP & {\bf French:} Ne pas utiliser chez l’ animal en gestation ou en période de lactation, \\
(Z=1.00) & car la sécurité du robenacoxib n’ a pas été établie chez les femelles gestantes ou \\
& allaitantes ni chez les chats et chiens utilisés pour la reproduction.\\ 
& {\bf English:} Do not use in pregnant or lactating animals because the safety of \\
& robenacoxib has not been established during pregnancy and lactation or in cats\\
& and dogs used for breeding.\\
\hline
\end{tabular}
\end{center}
\caption{This table shows for an example MTBT workload sentence from the EMEA French-English data how the optimal translation pair 
returned by MWNGP changes when going from Z = 0.00 to Z = 1.00. We provide the English translation of the MTBT workload sentence 
for the convenience of the reader since it was available from the EMEA parallel corpus. Note that in a real setting it 
would be the job of the translator to produce the English translation of the MTBT-French sentence using the 
translation pairs returned by MWNGP as help. }
\label{t:Z_ParamExample}
\end{table*}

Table~\ref{t:Z_ParamExample} shows an example of how the optimal 
translation pair returned by MWNGP changes from Z=0.00 to Z=1.00. The example illustrates the impact of changing the Z value on the nature of
the translation matches that get returned by MWNGP. As discussed in section~\ref{metrics}, smaller settings of Z are appropriate for
preferences for shorter matches that are more precise in the sense that a larger percentage of their content will be relevant. Larger
settings of Z are appropriate for preferences for longer matches that have higher recall in the sense that they will have more matches with
the content in the MTBT segment overall, although at the possible expense of having more irrelevant content as well. 

\section{Conclusions} \label{conclusions}

Translation memory is one of the most widely used translation technologies. 
One of the most important aspects of the technology is the system for assessing candidate translations from the TM bank for retrieval. 
Although detailed descriptions of the apparatus used in commercial systems are lacking, it is widely believed that they are based on 
an edit distance approach. 
We have defined and examined several TM retrieval approaches, including a new method using modified weighted n-gram precision that 
performs better than edit distance according to human translator judgments of helpfulness. 
The MWNGP method is based on the following premises: local context matching is desired; weighting words and phrases by expected
helpfulness to translators is desired; and allowing shorter n-gram precisions to contribute more to the final score than 
longer n-gram precisions is desired.
An advantage of the method is that it can be adjusted to suit translator length preferences of returned matches.
A limitation of MWNGP, and all other TM metrics we are aware of, is that they only consider the source language side.  
Examples from our experiments reveal that this can lead to poor retrievals. 
Therefore, future work is called for to examine the extent to which the target language sides of the translations in the TM bank influence 
TM system performance and to investigate ways to incorporate target language side information to improve TM system performance.

\bibliographystyle{acl}
\bibliography{paper}

\end{document}